# Naming Schema for a Human Brain-Scale Neural Network

Morgan Schaefer, Lauren Michelin, Jeremy Kepner
Massachusetts Institute of Technology

*Abstract-* **Deep neural networks have become increasingly large and sparse, allowing for the storage of large-scale neural networks with decreased costs of storage and computation. Storage of a neural network with as many connections as the human brain is possible with current versions of the high-performance Apache Accumulo database and the Distributed Dimensional Data Model (D4M) software. Neural networks of such large scale may be of particular interest to scientists within the human brain Connectome community. To aid in research and understanding of artificial neural networks that parallel existing neural networks like the brain, a naming schema can be developed to label groups of neurons in the artificial network that parallel those in the brain. Groups of artificial neurons are able to be specifically labeled in small regions for future study.**

## I. INTRODUCTION

As deep neural networks (DNNs) become larger and sparser, the opportunity to create a neural network with as many neurons and connections as the human brain becomes increasingly possible [1],[2]. To prepare for a neural network with over 100 billion neurons and over 100 trillion connections, a powerful database is needed [3],[4]. The high-performance Apache Accumulo database combined with D4M software allows for the storage and processing of a human brain-scale neural network [5].

Just as a human brain consists of connected neurons, a DNN consists of neurons that connect nodes, creating an output that is decided through a series of weights [6]. Larger DNNs become increasingly hard to store due to the amount of data they process; the best method for addressing this is to use sparse weight matrices [7]. Sparse weight matrices avoid storing weight values equal to zero, drastically decreasing the costs of storage and computation needed for large DNNs [8]. Seeing as most neurons in the human brain are connected to a small fraction of all the neurons, the human brain can be well represented by a sparse connection matrix.

Once a human brain-scale neural network is created and stored, it can be used to run various tests that may benefit the Connectome community [9]. In order for this neural network to be easily used and understood by neuroscientists and others for brain-related research, the neural network should be partitioned and labeled by groups of neurons in the human brain. Since the brain itself is sparse, it can be represented as a sparse matrix, thus taking. up less memory and allowing it to be easily stored as a sparse neural network, with each neuron in the network representing a neuron in the brain.

## II. APPROACH

This project continues a prior implementation of a large-scale neural network from 2018 [10]. Updates had to be made to the code so it could run on Julia v1.4.2 and Apache Accumulo v1.8. The code eventually aims for 100 trillion connections by creating and ingesting numerous linked $10^6$ by $10^6$ weight matrices with a sparsity of $10^{-3}$, with appropriate labels for the sections of neurons corresponding to different regions of the brain [10]. To test the efficiency and effectiveness of the updated code, various tests were run by ingesting and timing 50 million, 100 million, and 500 million entries using 1, 2, 4, 6, 8, 10, 12, 14, 16, and 18 database ingest processes. The number of entries was varied to test for consistency and capability with increasing load size, and the number of workers was varied to find the number that gives the most efficient ingest rate (see Figure 2).

A schema of existing, hierarchical brain regions was used to partition and label the neural network. For this project, the outer layer of the brain, where gray matter exists, was all that was considered, as gray matter contains the highest concentration of neurons [11] (see Figure 1). The two largest regions in this schema are the cerebral cortex and the cerebellum, both regions of gray matter, which contain the largest numbers of neurons in the brain [12]. The cerebral cortex has between 21 and 26 billion neurons, and it can be divided into 31 regions per hemisphere using available brain images and labels [12],[13]. From there, each region is divided into cortical columns, which are column-shaped regions of interconnected neurons [14]. The number of neurons in a cortical column varies throughout the cortex and throughout the neuroscience community; cortical columns are more conceptual regions than they are anatomical [15]. Some assert that there are between 1,000 and 10,000 neurons in a column [14]. Cortical columns can then be divided into cylindrical regions of about 100 neurons called microcolumns or minicolumns [16]. Due to the range of neurons in a cortical column, the number of microcolumns in a cortical column also varies (Table 1). Finally, each microcolumn is sliced horizontally by the five cell layers of the cerebral cortex. Although the cerebral cortex has six total layers, microcolumns only extend across layers II-VI, and layer I can be disregarded in this context due to neuron scarcity in the layer [16], [17]. The final region is a slice of the cylindrical microcolumn that consists of about 20 neurons (Figure 1).

TABLE 1
RANGES OF CEREBRAL CORTEX REGIONS

| Cortex Regions | Columns | Columns per cortex region | Microcolumns per column | Microcolumns | Layers | Microcolumns per cortex region | Regions (columns) | Neurons per region (columns) | Regions (microcolumns) | Neurons per region (microcolumns) | Total Neurons |
|---|---|---|---|---|---|---|---|---|---|---|---|
| 62 | 210,000 | 3,387 | 1,000 | 210,000,000 | 5 | 3,387,096 | 1,050,000 | 20,000 | 1,050,000,000 | 20 | 21,000,000,000 |
| 62 | 21,000,000 | 338,709 | 10 | 210,000,000 | 5 | 3,387,096 | 105,000,000 | 200 | 1,050,000,000 | 20 | 21,000,000,000 |
| 62 | 2,100,000 | 33,870 | 100 | 210,000,000 | 5 | 3,387,096 | 10,500,000 | 2,000 | 1,050,000,000 | 20 | 21,000,000,000 |
| 62 | 100,000 | 1,612 | 2,100 | 210,000,000 | 5 | 3,387,096 | 500,000 | 42,000 | 1,050,000,000 | 20 | 21,000,000,000 |
| 62 | 260,000 | 4,193 | 1,000 | 260,000,000 | 5 | 4,193,548 | 1,300,000 | 20,000 | 1,300,000,000 | 20 | 26,000,000,000 |
| 62 | 26,000,000 | 419,354 | 10 | 260,000,000 | 5 | 4,193,548 | 130,000,000 | 200 | 1,300,000,000 | 20 | 26,000,000,000 |
| 62 | 2,600,000 | 41,935 | 100 | 260,000,000 | 5 | 4,193,548 | 13,000,000 | 2,000 | 1,300,000,000 | 20 | 26,000,000,000 |
| 62 | 100,000 | 1,612 | 2,100 | 260,000,000 | 5 | 4,193,548 | 500,000 | 52,000 | 1,300,000,000 | 20 | 26,000,000,000 |

Table 1. This table shows possible numbers for different levels of the cerebral cortex naming schema, taking into account the minimum and maximum numbers of neurons in the cortex, and varying the amounts of columns per microcolumn (and subsequently neurons per column) based on assertions of 10, 100, and 1,000 microcolumns per column and 100,000 total columns [14]. Regardless of total neurons or microcolumns per column, the smallest final region contains 20 neurons.

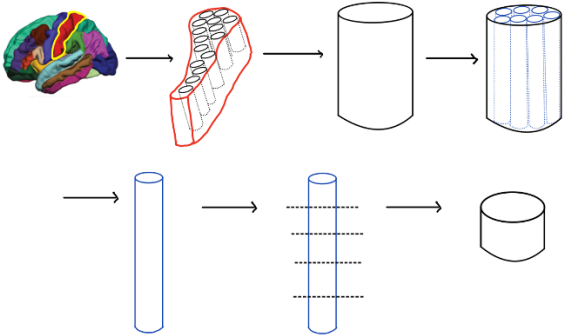

Fig 1. The organizational hierarchy for cerebral cortex brain region labels, from anatomical region to cortical column to microcolumn to final region.

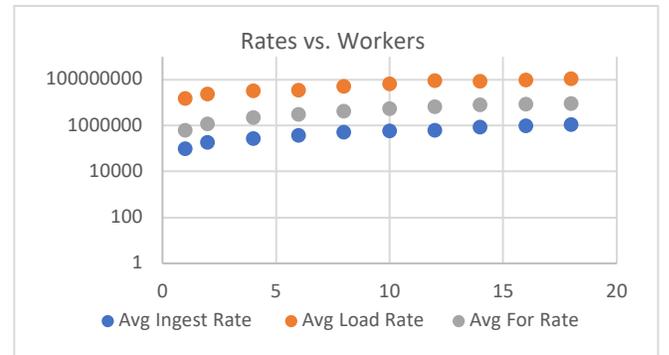

Fig 2. Average interest, load, and for rates per number of workers

The cerebellum, which has 100 billion neurons, can be partitioned similarly. There are three large functional regions that divide the cerebellum vertically, and there are 10 lobules that cut horizontally into each region [18], [19]. Further division is accomplished by five microzones which contain small groups of neurons called cerebellar modules [20], [21]. Modules are to the cerebellum as microcolumns are to the cerebral cortex: they are both the most basic units in each section of the brain [22]. Like cortical columns, there is variation in the number of neurons per module throughout the cerebellum [22]. As in the cerebral cortex, the final region is a module sliced by the three cell layers of the cerebellum [23].

The naming schema uses these regions to create labels for the smallest possible group of neurons in each section of the brain, thus generating the greatest possible number of regions. The list of regions is manipulated by the updated code to properly create, label, and ingest the weight matrices.

## III. RESULTS

After running tests with varying numbers of entries and workers, timing data showed, at numbers greater than eight workers, the timing to ingest a certain amount of data levels off (see Figure 2). It was also evident that ingest rate, load rate, and for rate increased with the number of workers. Ingest rate is the number of entries ingested per second, load rate is the number of entries loaded into the sparse weight matrices per second, and for rate is the number of entries labeled per second. The rates data shows that the load rate is not the bottleneck, and the limiting factor is the ingest rate. This indicates that, in a brain-scale neural network, the time for loading large sparse matrices would not hinder the ingest rate.

In labeling the weight matrices as regions of the brain using the schema described previously, over 1 billion regions can be generated and labeled for the cerebral cortex; if a minimum of 21 billion neurons is considered, there will be 1,050,000,000 regions, and if the maximum of 26 billion neurons is used, there will be 1,300,000,000 regions. The region names would appear as:

{hemisphere}/
  {anatomical region}/
    {assigned cortical column number between 1 and
    columns per anatomical region}/

{assigned microcolumn number between 1 and
    microcolumns per cortical column}/
     {layer}

The cerebellum follows a similar naming scheme and can be expected to have the same, if not greater, number of regions as the cerebral cortex due to its greater neuron density. The region names in the cerebellum would take the form:

{hemisphere}/
  {functional region}/
   {lobule}/
    {microzone}/
     {module}

There is less uniformity in neuron grouping throughout the cerebellum, and there is less agreement on numbers of neurons per basic operational structure; it is hard to establish uniform regions with as few as 20 neurons, as in the cerebral cortex, with the information and research currently available.

## IV. CONCLUSION

It should be possible to create a neural network with as many connections as the human brain with up-to-date code that utilizes D4M and the Accumulo database. To make this technology relevant and useful for the neuroscience and Connectome community, a naming schema has been proposed to liken neurons of the artificial network to neurons in the human brain. Over one billion regions can be labeled in the cerebral cortex, with even more regions in the cerebellum. This labeling schema aims to exist as a resource for understanding so that it is available when the Connectome community turns to artificial networks, which grow increasingly larger and sparser, for research.